\pdfoutput=1
\documentclass[journal,letterpaper]{IEEEtran}
\usepackage[letterpaper,left=0.775in,right=0.775in,top=0.75in,bottom=0.81in]{geometry}

\usepackage{amsmath,amsfonts}
\usepackage{algorithm}
\usepackage{algpseudocode}

\algtext*{EndIf} 
\algtext*{EndFor} 
\algrenewcommand\algorithmicindent{1.0em}
\usepackage{array}


\usepackage{textcomp}
\usepackage{stfloats}
\usepackage{url}
\usepackage{verbatim}
\usepackage{graphicx}   
\usepackage{cite}
\usepackage{subcaption}
\usepackage{enumitem}
\usepackage{booktabs}
\usepackage{tabularx}
\newcolumntype{C}{>{\centering\arraybackslash}X}
\usepackage{multirow}
\usepackage{makecell}
\usepackage{wrapfig}
\usepackage{xcolor}
\usepackage[hidelinks]{hyperref}

\newcounter{expquestion}

\newcommand{\question}[1]{%
    \refstepcounter{expquestion}%
    \subsection*{Q\theexpquestion: #1}%
}

\begin{document}
\pagestyle{empty}

\title{RoboPrompt: Intuitive Robot Policy Steering with Sparse Human Input}

\author{Yanwen Zou$^{1,2}$, Chenyang Shi$^{1}$, Guoxuan Xu$^{1}$, Wenye Yu$^{1,2}$,\\
Wendi Chen$^{1,2}$, Ye Pan$^{1}$, Cewu Lu$^{1,2,3,\dagger}$, Chuan Wen$^{1,\dagger}$\\[0.5ex]
{\normalsize $^{1}$Shanghai Jiao Tong University\quad
$^{2}$Shanghai Innovation Institute\quad $^{3}$Noematrix Ltd.}\\[0.5ex]
{\normalsize \url{https:/yanwen-zou.github.io/Roboprompt-Website/}}%
\thanks{$^{\dagger}$Corresponding authors.}%
}


\maketitle
\thispagestyle{empty}

\begin{abstract}
End-to-end robot policies trained through imitation learning remain constrained by limited data diversity, making reliable zero-shot deployment in real-world settings challenging. Shared-autonomy methods enable human correction through teleoperation, but specialized hardware and operator training hinder deployment at scale. Other approaches incorporate human guidance as additional policy inputs, often requiring architectural changes and dedicated training for steerability, which limits their applicability across policies. We present RoboPrompt, a general-purpose, lightweight robot policy steering system that enables users to guide policy behavior through intuitive, sparse inputs, including drawn traces, target points, and coarse directional instructions. RoboPrompt decouples human-intention translation from the underlying policy: a reusable module converts human guidance into action drafts, which are refined through the diffusion or flow-matching dynamics of the base policy. By controlling action generation in noise space, RoboPrompt balances human intent with the policy prior without modifying the base policy architecture or fine-tuning it for steerability. Experiments demonstrate effective steering across Diffusion Policy, $\pi_{0.5}$, and FastWAM. We further use steered rollouts for online policy improvement through DAgger. After 2-3 rounds of iteration, average success rates increase by 15.5\% for $\pi_{0.5}$ across three tasks and by 21.3\% across three policies(Diffusion Policy, $\pi_{0.5}$, FastWAM) on the Insert Bread task, while average human intervention counts decrease by 44.0\% (2.86 to 1.60) and 81.9\% (2.60 to 0.47), respectively.
\end{abstract}

\section{Introduction}

\begin{figure}[t]
    \centering
    \includegraphics[width=0.96\columnwidth]{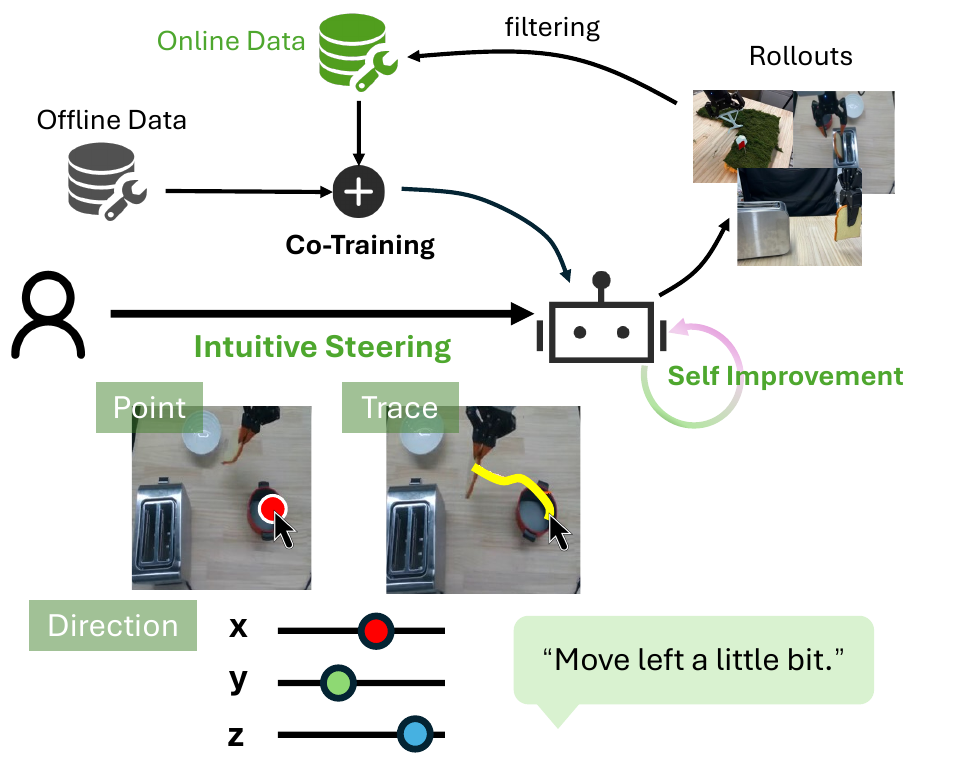}
    \caption{\textbf{Overview of RoboPrompt}. RoboPrompt aims to enable human-in-the-loop deployment of robot policies, collecting failure-recovery rollout data that can drive continual policy improvement. It supports intuitive multimodal steering through inputs, including points, traces, and coarse directional instructions. These human-guided rollouts are collected and combined with offline data for co-training, forming a closed-loop framework for task completion and self-improvement without teleoperation.}
    \label{fig:overview}
\end{figure}

\IEEEPARstart{R}{ecent} advances in vision-language-action (VLA) models\cite{intelligence2025pi05visionlanguageactionmodelopenworld,brohan2022rt,kim2024openvla,zitkovich2023rt} and world-action-models(WAM)\cite{yuan2026fast,ye2026world,li2026causal,zhang2026native} have significantly improved the generalization capability of robotic manipulation.

Scaling beyond laboratory settings remains constrained by scarce real-world interaction data\cite{goldberg2025good}, unlike the web-scale corpora available to language and vision models\cite{liu2024deepseek,team2025kimi}. This creates a chicken-and-egg dilemma: limited data diversity causes deployment failures, which in turn restrict deployment and further data collection.

\textbf{Human-in-the-loop deployment} addresses this dilemma by letting robots act autonomously while humans monitor and correct execution. However, end-to-end policies\cite{intelligence2026pi,kim2024openvla,intelligence2025pi,chi2025diffusion,yuan2026fast} offer limited intervention interfaces. Language specifies task objectives but often provides weak fine-grained control\cite{glossop2026cast,fei2025depth}, especially when the intended behavior is semantically ambiguous\cite{yu2025point,huang2026graphcot}.

Shared-autonomy approaches allow humans to correct failures through teleoperation\cite{intelligence2025pi,pan2026sop,yu2026armada}, but require additional hardware and operator training, hindering adoption by everyday users after large scale deployment.

We argue that \textbf{scalable human-in-the-loop deployment requires an intuitive steering interface with minimal physical and cognitive burden.} Drawing 2D sketches on camera observations or providing coarse directional instructions could enable task completion while collecting corrective data for policy improvement.

Realizing such an interface presents several challenges. First, unlike teleoperation, natural human instructions only provide partial information rather than full 6D actions with temporal dynamics. Second, long-horizon prompt following is non-trivial under action-chunked policies. Effective correction may require actions outside the base policy's learned action manifold, which the policy cannot reliably produce on its own, so steering should initially prioritize human intention and gradually transition toward the policy prior across subsequent action chunks. Finally, existing prompt-conditioned approaches often rely on whole policy fine-tuning or architectural modifications\cite{chen2026steerable,yu2025point,zheng2025tracevla}, which ties steerability to a particular policy design and makes it difficult to reuse the same steering interface across pretrained policies with different architectures or weights.

We propose \textbf{RoboPrompt}, a two-stage system that translates sparse human input into executable actions, integrates them with pretrained policies, and uses steered rollouts for policy improvement. Our main contributions are:

\begin{itemize}
    \item \textbf{From intention to action draft.}
    We introduce a lightweight human-intention translator based on a 0.77B VLA model\cite{lin2026evo}, together with a data processing pipeline for its training. The translator converts 2D traces, points, and coarse directional instructions into spatially and temporally complete action chunks, serving as a task-agnostic yet physically plausible action generator.

    \item \textbf{From action draft to fine-grained movement.}
    We propose \textit{Progressive Biased Denoising (PBD)}, which refines action drafts by exploiting the properties of diffusion- and flow-matching-based policies. Increasing the denoising level across action chunks extends a prompt's influence over long-horizon behavior while smoothly shifting control toward the policy prior.

    \item \textbf{Steering and improving real-world policies.} RoboPrompt achieves $\geq95\%$ prompt alignment on Diffusion Policy\cite{chi2025diffusion}, $\pi_{0.5}$\cite{intelligence2025pi}, and FastWAM\cite{yuan2026fast} without fine-tuning for steerability. We introduce Time-weighted Optimal Transport (TOT) to filter steered rollouts for DAgger training, improving task progress across real-world tasks while reducing human intervention.
\end{itemize}

\section{Related Work}

\textbf{Robot learning for manipulation tasks.} 
Learning-based visuomotor policies, ranging from task-specific robot policies\cite{chi2025diffusion} to large-scale robot foundation models including VLAs\cite{intelligence2025pi,intelligence2026pi} and WAMs\cite{li2026causal,yuan2026fast}, have achieved increasingly strong capabilities in robotic manipulation through imitation learning. These policies are typically trained end to end to map visual observations and task instructions directly to robot actions, allowing them to benefit from large-scale robot datasets. However, this end-to-end formulation provides few interfaces for humans to intervene during policy execution.


\textbf{Policy steering.} 
To improve manipulation performance, recent works have explored steering a base policy with additional conditioning signals during execution. Steerable Policies\cite{chen2026steerable} enable users to provide diverse prompts for refining policy behavior by translating them into language inputs to condition the policy. $\pi_{0.7}$\cite{intelligence2026pi} takes a step further toward more grounded conditioning by prompting a world model to convert subtask-level language prompts into goal images. While language is effective for specifying high-level tasks and semantic intent, language alone often cannot express the full set of fine-grained spatial and motion-level information required for effective steering\cite{yu2025point,wang2026vpvlavisualpromptinginterface}. Inference-Time Policy Steering\cite{wang2025inference} enables training-free intervention by using visual guidance objectives to bias the sampling process of diffusion policy\cite{chi2025diffusion}. However, reweighting remains constrained by the base policy's own action distribution, limiting its ability to realize corrective actions outside the policy's learned action manifold that may be necessary for effective failure recovery. Instead, RoboPrompt uses an independent module to convert human input into an action draft and refines it using Progressive Biased Denoising (PBD) and Noise-Augmented Flow Reversal Steering (NA-FRS) with a tunable initial denoising step, preserving human intent while adjusting its influence relative to the base policy prior.

\textbf{Interactive robot learning.} 
Interactive robot learning enables a learned policy to receive human guidance or intervention during execution. Such interactions improve task completion when autonomous execution is unreliable, and the resulting trajectories can further serve as online data for policy improvement methods such as HG-DAgger\cite{kelly2019hg}. Existing collaborative-control systems\cite{liu2025robot, wu2025robocopilot, yu2026armada} commonly realize this interaction through teleoperation, where users directly provide low-level motion commands. Shared-autonomy methods such as HAJL\cite{luo2025human} reduce the control burden by employing a diffusion-based assistive policy that refines teleoperation commands during the diffusion process, continuously interpolating between manual and autonomous control. Nevertheless, these approaches still depend on teleoperation devices and expert operators to produce precise low-level commands. To reduce this dependence on skilled teleoperation, RoboPrompt uses sparse, intuitive human guidance and relies on the base policy as an action prior to infer executable low-level behavior. 

\section{System Architecture}
Human intentions often underspecify robot actions. We seek a general mapping from sparse human guidance to action drafts that downstream policies can refine using their task-aware priors. RoboPrompt implements this mapping in two phases. \textbf{Phase I} uses a lightweight steer model to translate points, traces, and coarse directional instructions into continuous action drafts. \textbf{Phase II} refines these drafts through the downstream policy's denoising process, balancing human intent with the policy prior without policy fine-tuning or architectural changes. We detail these two phases in the following subsections.
\subsection{Phase I: Action Drafting}
\begin{figure}[!t]
    \centering
    \includegraphics[width=0.88\columnwidth]{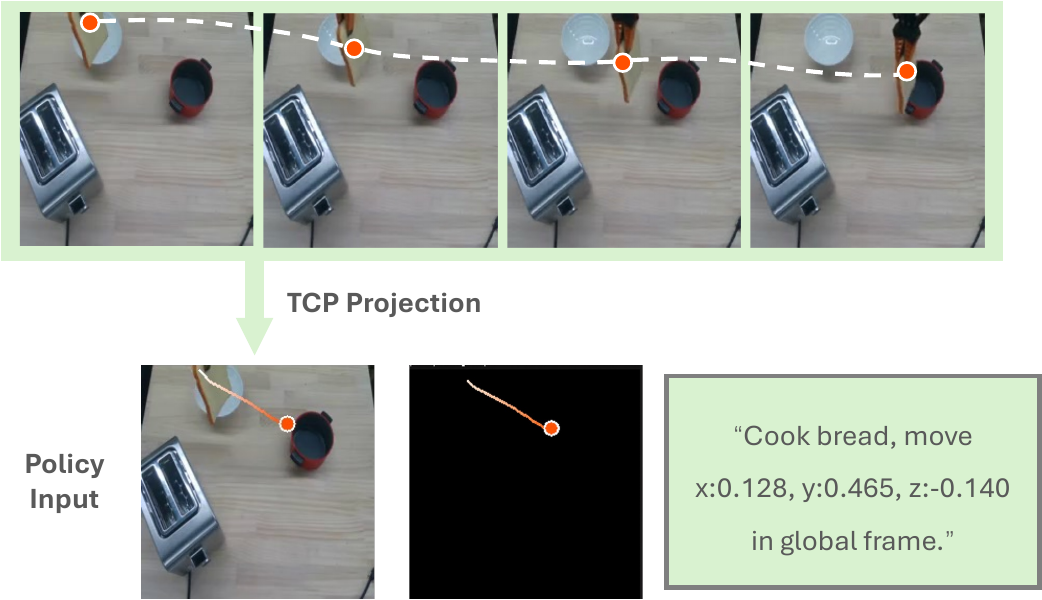}
    \caption{\textbf{Prompt label generation.}
    Over the next n frames, connected TCP projections and the final TCP projection provide ground truth for trajectory and point prompts; summed and normalized actions supervise the global action prompt.}
    \label{fig:evo1_input}
\end{figure}

Intuitively, humans often teach others how to perform a task by drawing a sketch, pointing to a desired location, or giving a coarse directional instruction. RoboPrompt aims to provide robots with the same lightweight forms of guidance. We therefore define three prompt types: point and trajectory prompts are given as visual inputs directly overlaid on the camera observation, while global action prompts are given as text inputs appended to the task instruction. Though VLMs can produce similar prompt labels, we find that they can be unreliable when the task contains semantic ambiguity, visual occlusion, or fine-grained manipulation that requires precise motion reasoning. We therefore implement prompt label generation as follows. For each trajectory, we construct the prompt that would have guided the robot along its actually executed future motion, as illustrated in Fig.~\ref{fig:evo1_input}.

\begin{figure*}[t]
    \centering
    \includegraphics[width=0.96\textwidth]{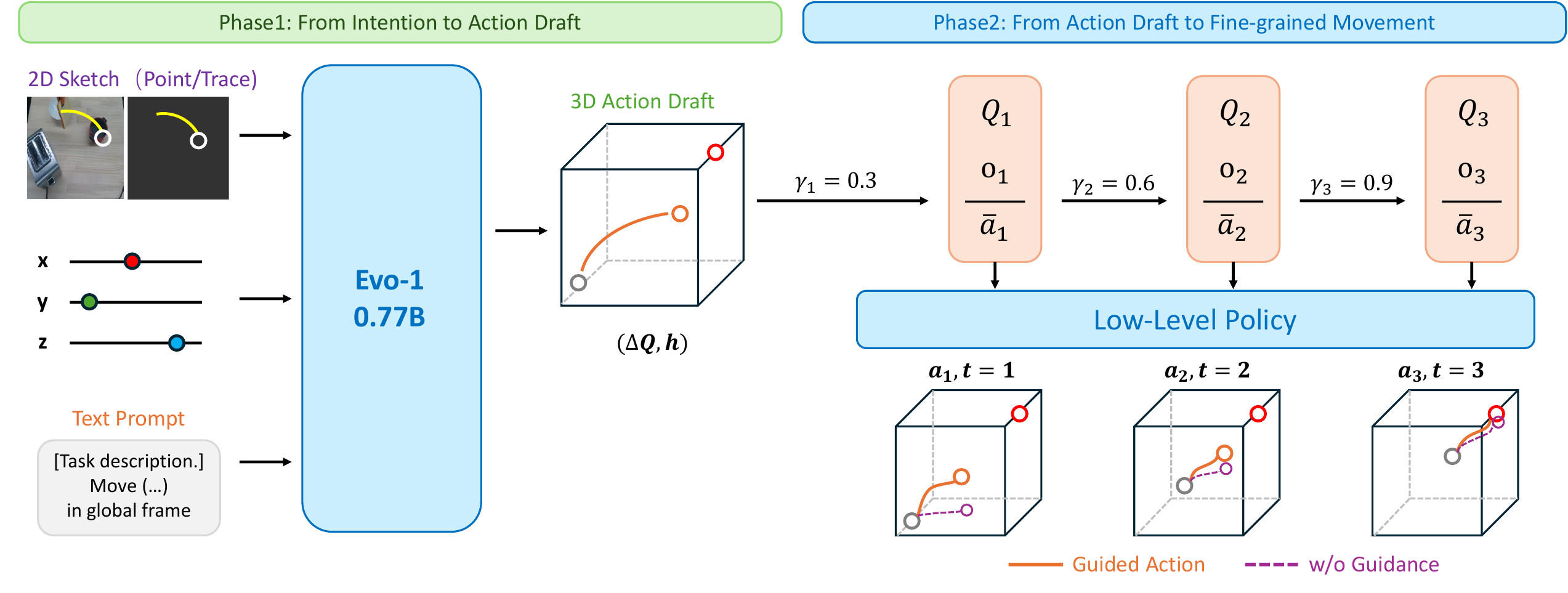}
    \caption{\textbf{System Architecture.}
    RoboPrompt incorporates human intention into policy inference through two phases.
    In Phase I, prompt inputs are processed by a lightweight VLA to produce a temporally continuous 6D action chunk as an action draft.
    In Phase II, the base policy refines the action draft generated by Phase I. The fraction of noise-adding steps, $\gamma$, controls the balance between human guidance and the base policy. $\gamma$ increases with successive replanning steps, gradually shifting control toward the base policy.}
    \label{fig:architecture}
\end{figure*}

Formally, let an episode be $\tau=\{(O_t,Q_t,a_t)\}_{t=0}^{T-1}$, where $O_t$ is the visual observation, $Q_t$ is the robot TCP pose, and $a_t\in\mathbb{R}^{d_a}$ is the low-level action. Denote the action horizon as $H$. We use an image projection function
\begin{equation*}
    u_t = f_{\mathrm{proj}}(Q_t) \in \mathbb{R}^2 ,
\end{equation*}
to project the TCP onto the image plane of the robot's fixed environment camera.

\textbf{Point prompt.}
A point prompt specifies where the TCP should arrive after the short-horizon motion. For a sample starting at time $t$, the hindsight target is the projected TCP endpoint
\begin{equation*}
    p_t^{\mathrm{pt}} = u_{\bar{t}}, \qquad
    \bar{t}=\min(t+H-1,T-1).
\end{equation*}
The model input is an image with $p_t^{\mathrm{pt}}$ drawn on top of $O_t$,
\begin{equation*}
    \tilde{O}_t^{\mathrm{pt}} =
    \mathrm{Overlay}_{\mathrm{pt}}(O_t,p_t^{\mathrm{pt}}).
\end{equation*}

\textbf{Trajectory prompt.}
A trajectory prompt gives a sparse 2D trace of the desired TCP motion over the action horizon. Its ground truth is the projected future TCP sequence
\begin{equation*}
    P_t^{\mathrm{traj}}
    =
    \left(u_t,u_{t+1},\ldots,u_{\bar{t}}\right),
    \qquad
    \bar{t}=\min(t+H-1,T-1),
\end{equation*}
and the visual input is produced by connecting these sparse points into a line on the observation,
\begin{equation*}
    \tilde{O}_t^{\mathrm{traj}} =
    \mathrm{Overlay}_{\mathrm{traj}}(O_t,P_t^{\mathrm{traj}}).
\end{equation*}

\textbf{Global action prompt.}
A global action prompt describes the coarse Cartesian displacement that the robot should make in the base frame. Its hindsight target is obtained by accumulating the raw XYZ actions over the same horizon,
\begin{equation*}
    g_t = \mathrm{Norm}_{g}\!\left(\sum_{k=0}^{H-1} a_{t+k}^{xyz}\right)\in[-1,1]^3 .
\end{equation*}
It is then converted into a textual prompt, as shown in Fig.~\ref{fig:evo1_input}. For all three prompts, we normalize the input and apply random position and scale perturbations during training, considering that humans always provide imprecise prompts.

Phase I adopts a lightweight 0.77B-parameter VLA, Evo-1~\cite{lin2026evo}, as the action-draft policy architecture, as shown in Fig.~\ref{fig:architecture}. For point and trajectory prompts, the prompt is rendered directly on the environment-camera image and used as the primary visual input. To make the steering signal explicit and prevent it from being diluted by scene appearance, we also render a prompt-only image without the camera background and feed it as the second image input. For global action prompts, the normalized Cartesian displacement is verbalized and appended to the original task description as part of the text input. The model predicts a temporally continuous action chunk,
\begin{equation*}
    a_{\mathrm{I}} = \pi_{\phi}(O_t^{(1)}, O_t^{(2)}, q_t),
\end{equation*}
where $O_t^{(1)}$ is the prompted camera observation, $O_t^{(2)}$ is the prompt-only image for visual prompts, and $q_t$ is the task instruction augmented with the global action prompt when applicable.

In practice, we train the Phase I steering model on 500 play-data trajectories collected on the same robot platform used in our experiments, with prompt labels generated using the procedure described above. Auxiliary losses are applied to supervise the training process as follows:
\begin{equation*}
    \mathcal{L}_{\mathrm{I}} =
    \mathcal{L}_{\mathrm{act}}(a_{\mathrm{I}}, a_{t:t+H-1})
    + \lambda_{\mathrm{aux}}\mathcal{L}_{\mathrm{aux}}.
\end{equation*}
The auxiliary objective combines supervision for the three prompt types,
\begin{equation*}
    \mathcal{L}_{\mathrm{aux}} =
    \mathcal{L}_{\mathrm{pt}}
    + \mathcal{L}_{\mathrm{traj}}
    + \mathcal{L}_{\mathrm{global}},
\end{equation*}
The point loss $\mathcal{L}_{\mathrm{pt}}$ and trajectory loss $\mathcal{L}_{\mathrm{traj}}$ supervise the target point and trajectory at the pixel level in the visual input space. For global action prompts, we use $\mathcal{L}_{\mathrm{global}}=\operatorname{MSE}(\hat{g}_t,g_t)$, where $\hat{g}_t$ is the predicted global action and $g_t$ is the normalized ground-truth displacement defined above. At this stage, the model produces plausible actions consistent with human intent as well as current observation. The resulting action draft is further refined in Phase II.

\subsection{Phase II: From Action Draft to Fine-grained Movement}
Phase II aims to refine the action draft produced by Phase I with the base policy prior. Current mainstream robot policies commonly generate actions through diffusion or flow-matching processes. Following \cite{luo2025human}, we leverage the mathematical structure of iterative denoising to fuse human intention with the policy output. To enable long-horizon prompts to influence policy behavior across action chunks while smoothly transitioning control from human intention toward the policy prior, we propose Progressive Biased Denoising (PBD). Let $m$ denote the replanning chunk index after a prompt is issued, $N$ be the total denoising steps of the base policy, and $k^m$ be the current denoising step, updated by a fixed increment $\Delta k$:
\begin{equation*}
    k^m=\min(k^0+m\Delta k,N),\qquad \gamma^m=k^m/N.
\end{equation*}
Here $\gamma^m$ is the chunk-wise normalized denoising level used for refinement. In the basic PBD formulation, we perturb the Phase I draft at each chunk with the progressively increasing $\gamma^m$ and denoise it with the base policy:
\begin{equation*}
    z_{\gamma^m}=(1-\gamma^m)a_{\mathrm{I}}+\gamma^m\epsilon,\qquad \epsilon\sim\mathcal{N}(0,I),
\end{equation*}
\begin{equation*}
    a_{\mathrm{II}}=\mathcal{D}_{\theta}(z_{\gamma^m},\gamma^m\rightarrow0),
\end{equation*}
where $z_{\gamma^m}$ denotes the noisy action at denoising level $\gamma^m$, and $\mathcal{D}_{\theta}$ denotes the denoising process of the downstream policy. Note that $\gamma$ is a tunable control ratio that regulates the influence of human guidance relative to the policy's original prediction. When $\gamma=0$, the Phase II output is exactly the Phase I draft, while when $\gamma=1$, the Phase II output falls back to the base policy action without any human guidance.

A single human sketch or directional prompt often corresponds to motion longer than one action chunk. The monotonic schedule of $\gamma^m$ preserves the same Phase I human input across multiple replanning chunks: the first chunk follows the human input strongly, while later chunks gradually reintroduce the base policy prior.

However, the action change induced by denoising is not uniform across denoising time: even a small number of denoising steps can produce strong corrections, as further illustrated in our denoising-dynamics experiment in Sec.~\ref{sec:denoise_dynamics}. This is undesirable for cross-action-chunk prompting, where control should transition smoothly rather than abruptly from human intention to the policy prior. To avoid the strong correction caused by directly adding random noise to $a_{\mathrm{I}}$, we build on Flow Reversal Steering\cite{tang2026improving} and introduce Noise-Augmented Flow Reversal Steering (NA-FRS). NA-FRS first runs the learned denoising dynamics backward from the clean Phase I action to the target denoising level as FRS:
\begin{equation*}
    z_{\gamma^m}^{\mathrm{rev}}=\mathcal{R}_{\theta}(a_{\mathrm{I}},0\rightarrow \gamma^m),
\end{equation*}
where $\mathcal{R}_{\theta}$ is the reverse integration of the base policy denoising dynamics. We write the implied reverse noise as
\begin{equation*}
    \epsilon_{\mathrm{rev}}=(z_{\gamma^m}^{\mathrm{rev}}-(1-\gamma^m)a_{\mathrm{I}})/\gamma^m.
\end{equation*}
This reverse process largely preserves Phase I information while limiting the original policy performance. So we gradually corrupt the Phase I distribution as $\gamma^m$ grows. NA-FRS therefore mixes the reverse path with pure Gaussian noise:
\begin{equation*}
    \epsilon_{\mathrm{mix}}=\sqrt{1-\rho^2}\epsilon_{\mathrm{rev}}+\rho\epsilon,
\end{equation*}
\begin{equation*}
    z_{\gamma^m}^{\mathrm{NA}}=(1-\gamma^m)a_{\mathrm{I}}+\gamma^m\epsilon_{\mathrm{mix}},
\end{equation*}
where $\rho\in[0,1]$ controls the random-noise ratio. The refined action is then obtained by denoising $z_{\gamma^m}^{\mathrm{NA}}$ back to $t=0$.
\subsection{Improve Policy Performance with Steered Rollouts}
Successful steered rollouts may still contain oscillations, repeated gripper actions, or long suboptimal paths. Such data can pair similar observations with inconsistent or delayed actions, degrading Markovian imitation-learning policies during DAgger-style\cite{ross2011reduction} training. We introduce Time-weighted Optimal Transport (TOT) to retain successful rollouts whose behavior and duration are close to expert demonstrations.

Concretely, for a steered rollout $\tau^r=\{x_i^r\}_{i=1}^{T_r}$ and an expert trajectory $\tau^e=\{x_j^e\}_{j=1}^{T_e}$, we extract policy embeddings $\phi(x)$ and define a timestep-level cost
\begin{equation*}
    C_{ij}=1-\frac{\phi(x_i^r)^\top \phi(x_j^e)}
    {\|\phi(x_i^r)\|_2\|\phi(x_j^e)\|_2}.
\end{equation*}
We then compute a TOT score by matching the rollout to its closest expert trajectory while penalizing unnecessarily long executions. Let $\bar{T}_E=\mathrm{median}_{\tau\in\mathcal{D}_E}|\tau|$:
\begin{equation*}
    D_{\mathrm{OT}}(\tau^r,\tau^e)=\min_{\Pi\in\mathcal{U}(T_r,T_e)}\sum_{i=1}^{T_r}\sum_{j=1}^{T_e}\Pi_{ij}C_{ij}.
\end{equation*}
\begin{equation*}
    S(\tau^r)=\min_{\tau^e\in\mathcal{D}_E}\left[D_{\mathrm{OT}}(\tau^r,\tau^e)+\lambda\max\!\left(0,T_r/\bar{T}_E-1\right)\right].
\end{equation*}
where $D_{\mathrm{OT}}$ denotes the optimal-transport distance between a rollout and an expert trajectory, $\Pi$ is the transport plan between rollout and expert timesteps, and $\mathcal{U}(T_r,T_e)$ denotes the set of plans with uniform temporal marginals. Lower scores indicate rollouts that are both semantically close to expert behavior and not excessively long. We only keep rollouts with $S(\tau^r)$ below a threshold for online training. Following \cite{fang2026robopocket}, the filtered online rollouts are mixed with the original offline data at a $50\%/50\%$ ratio for the next DAgger training round, gradually moving the original policy distribution toward the corrected deployment distribution while retaining the base policy prior.

\section{Experiment}
\begin{table*}[t]
\centering
\caption{Steering performance of RoboPrompt across different policies and tasks. The left block compares task progress with and without steering; the right block reports steering quality (the lower the steering counts, the less human intervention is needed; the higher the prompt alignment, the more faithfully the steered behavior follows the human prompt).}
\label{tab:steering_results}
\resizebox{\textwidth}{!}{
\begin{tabular}{llcc|cccc}
\toprule
\multicolumn{4}{c}{\textbf{Success Rate Improvement}} &
\multicolumn{4}{c}{\textbf{Steering Quality}} \\
\cmidrule(r){1-4} \cmidrule(l){5-8}
Policy & Task
& \makecell{Progress without\\Steering}
& \makecell{Progress with\\Steering}
& \makecell{Avg. Steering\\Counts}
& \makecell{Avg. visual\\input}
& \makecell{Avg. global\\input}
& \makecell{Avg. Prompt\\Alignment (\%)} \\
\midrule

\multirow{3}{*}{$\pi_{0.5}$}
& Left  & \multirow{3}{*}{59.6\%} & 77.5\% & 2.50 & 1.45 & 1.90 & \multirow{3}{*}{99.44\%} \\
& Right & & 72.4\% & 2.95 & 2.50 & 2.50 & \\
& Pot   & & 89.8\% & 2.35 & 1.80 & 1.25 & \\
\midrule

\multirow{3}{*}{DP}
& Left  & \multirow{3}{*}{46.5\%} & 71.3\% & 4.80 & 3.10 & 3.40 & \multirow{3}{*}{97.1\%} \\
& Right & & 64.75\% & 3.45 & 1.85 & 1.95 & \\
& Pot   & & 57.95\% & 4.40 & 3.95 & 1.85 & \\
\midrule

\multirow{3}{*}{FastWAM}
& Left  & \multirow{3}{*}{46.2\%} & 57.4\% & 3.15 & 2.15 & 2.30 & \multirow{3}{*}{100.0\%} \\
& Right & & 62.3\% & 3.10 & 1.90 & 2.10 & \\
& Pot   & & 52.3\% & 2.65 & 1.70 & 1.55 & \\
\bottomrule
\end{tabular}
}
\end{table*}

We use a Flexiv Rizon4 robot equipped with a Robotiq 2F-85 gripper as our robot platform. For each task, we collect 100 demonstrations through teleoperation with a Sigma.7 device to train the base policy.
\setcounter{expquestion}{0}

\question{Can RoboPrompt Effectively Steer Policies toward Different Task Objectives?}

\begin{figure}[t]
    \centering
    \IfFileExists{pics/steer_setup_with_pies.pdf}{%
        \includegraphics[width=0.48\textwidth]{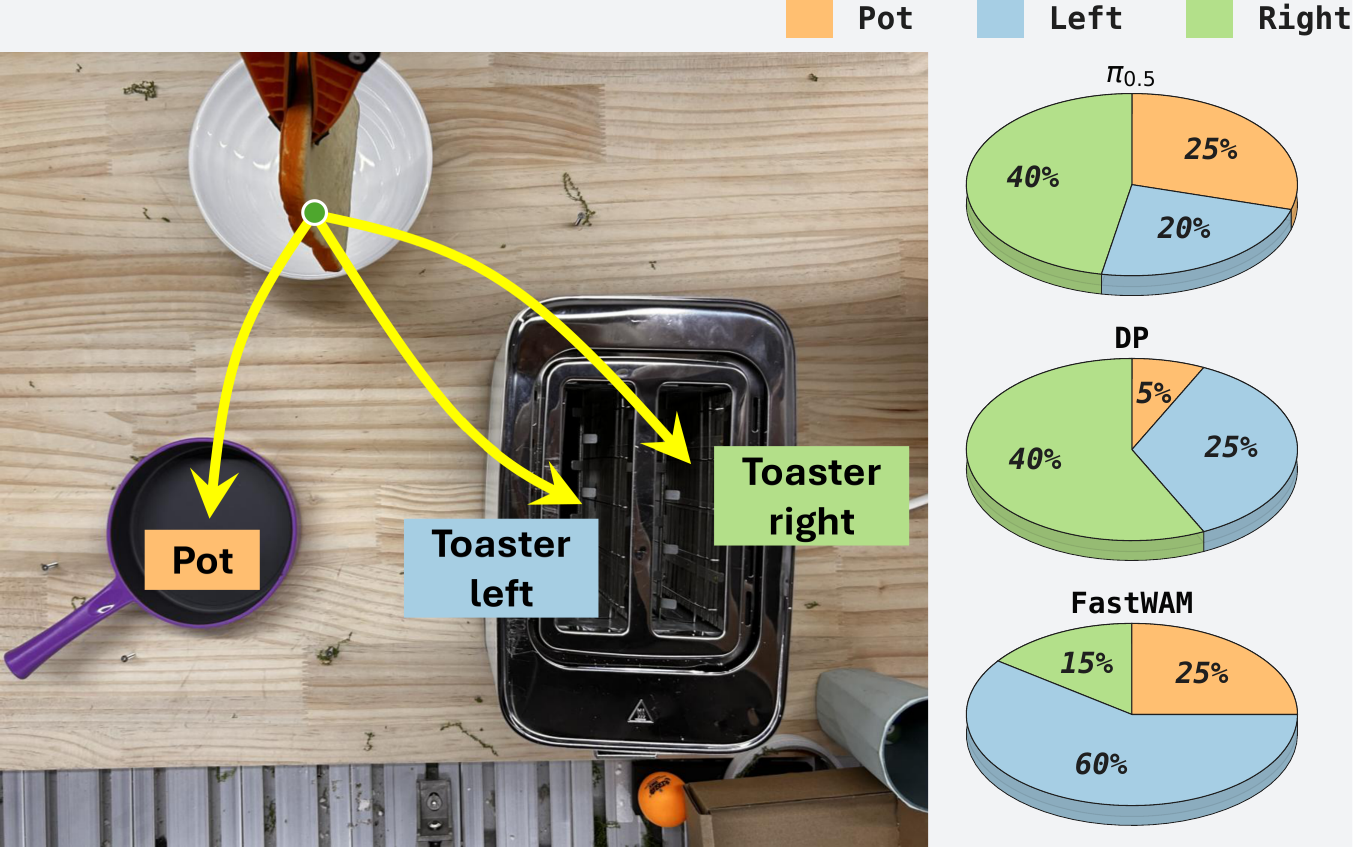}%
    }{%
        \fbox{\parbox[c][0.35\textheight][c]{0.44\textwidth}{\centering Missing figure:\\pics/steer\_setup\_with\_pies.pdf}}%
    }
    \caption{\textbf{Steering experiment setup and natural policy distribution.}
    In the steering experiment, the robot picks up bread from the bowl and places it into one of three targets (pot, left toaster, right toaster).
    The training set contains 46 demonstrations per target, all using the same text description across the three targets.
    The right panel shows the natural distribution of the three base policies over the three targets before steering.}
    \label{fig:steer_setup}
\end{figure}

We conduct steering experiments on Diffusion Policy (DP), $\pi_{0.5}$, and FastWAM. For each policy, we train on data containing three different task objectives, resulting in a multimodal policy distribution. RoboPrompt is then used at inference time to specify the desired task objective, as illustrated in Fig.~\ref{fig:steer_setup}.

Without steering, the learned multimodal policies exhibit different preferences over the three valid task objectives, as shown by the pie charts in Fig.~\ref{fig:steer_setup}. Note that the percentages for a given policy do not sum to $100\%$ because some trials failed before reaching any target during inference; those failures are excluded from the task-preference distribution.

We then apply RoboPrompt to steer each policy toward each specific target. We define prompt alignment as the percentage of human interventions for which the resulting robot motion follows the direction specified by the prompt along the dimensions indicated by that input. Dimensions not specified by the prompt are excluded from this assessment. For each policy, we compute this percentage over all interventions across the three task objectives. The results are reported in Table~\ref{tab:steering_results}, which is divided into two blocks. The left block compares the task progress with and without steering: across all three policies, RoboPrompt consistently improves the success rate over the unsteered baseline. The right block evaluates the steering quality: prompt alignment remains above $95\%$ across all three policies, confirming that the steered actions follow the human prompt.

\question{Can RoboPrompt Help Policy Improvement via DAgger?}

\begin{figure*}[t]
    \centering
    \begin{minipage}[b]{0.58\textwidth}
        \centering
        \includegraphics[width=\linewidth]{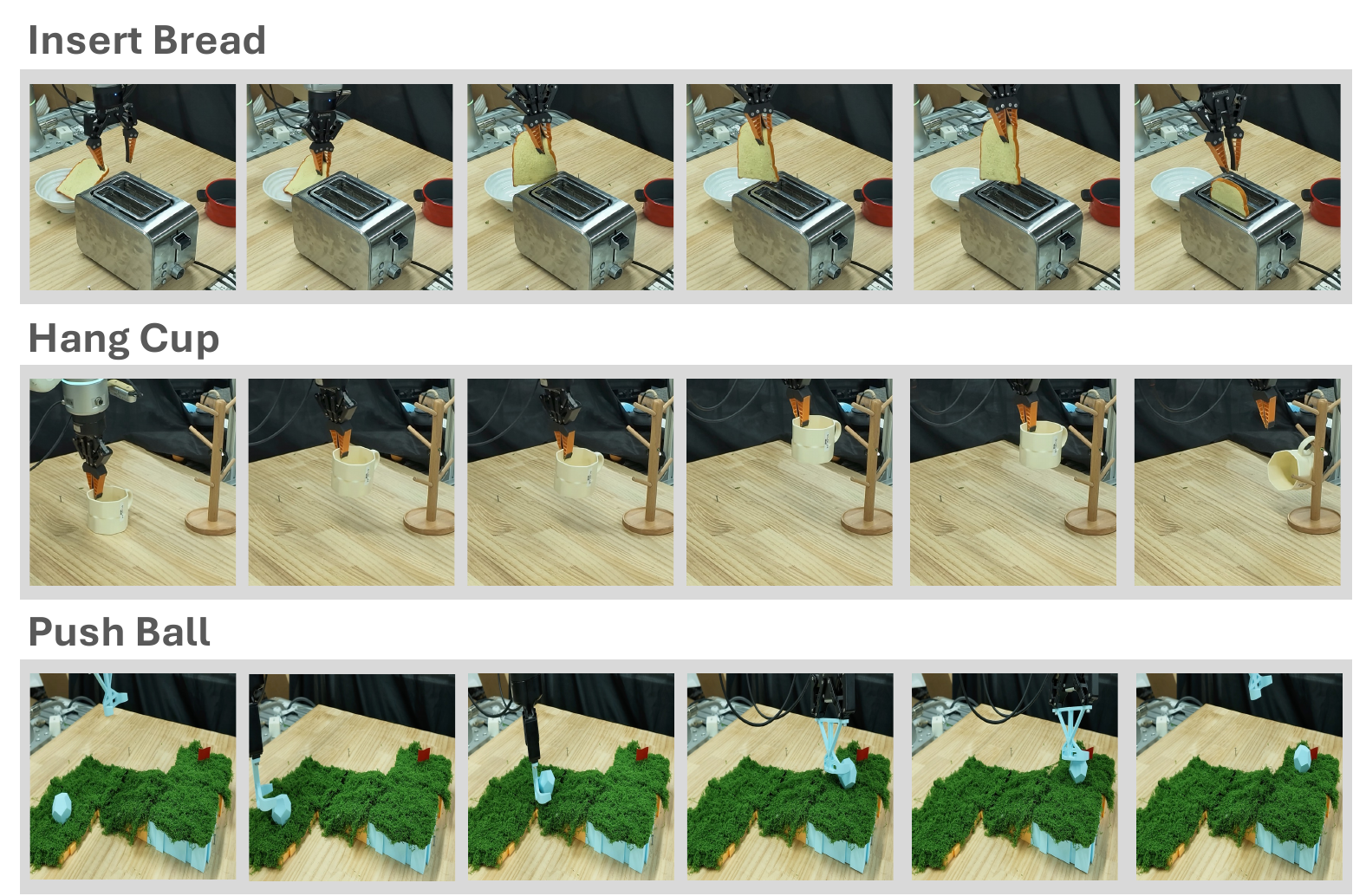}
        \par\smallskip
        {\small (a)}
    \end{minipage}\hspace{0.02\textwidth}
    \begin{minipage}[b]{0.26\textwidth}
        \centering
        \includegraphics[width=\linewidth]{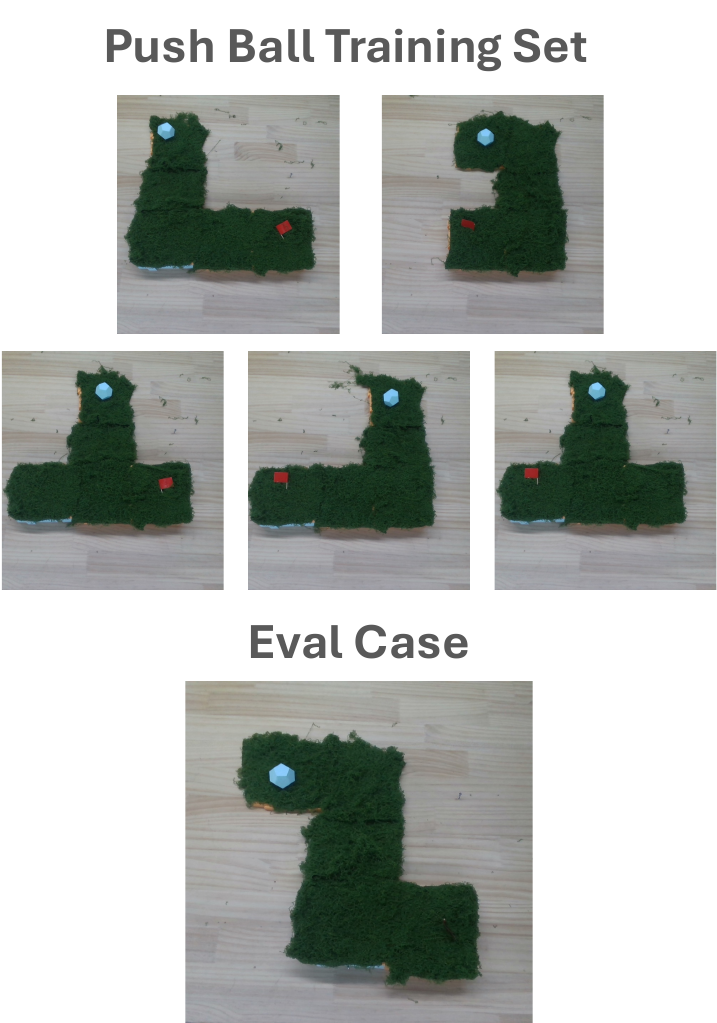}
        \par\smallskip
        {\small (b)}
    \end{minipage}
    \caption{\textbf{DAgger experiment setup.}
    (a) \textit{Insert Bread}: pick up bread from a bowl and insert it into the toaster's right slot; \textit{Hang Cup}: hang a cup by its handle on the second-highest peg; \textit{Push Ball}: push a polyhedral ball across an uneven platform until it touches the flag.
    (b) Push Ball training dataset contains 5 different layouts, and is evaluated on another new layout. Steered rollouts enable task completion in the new layout and provide data improve the policy over successive DAgger rounds.}
    \label{fig:dagger_setup}
\end{figure*}

Based on the filtered steered-rollout pipeline described above, we evaluate whether RoboPrompt can improve the underlying policy through DAgger. We first evaluate this with $\pi_{0.5}$ on three real-world tasks: Insert Bread, Hang Cup, and Push Ball (Fig.~\ref{fig:dagger_setup}). Each task uses 100 training demonstrations: Insert Bread uses only right-slot insertion demonstrations, unlike the insertion task in the steering experiment, and Push Ball uses 20 demonstrations for each of five different routes. At inference, Push Ball uses an unseen layout to evaluate adaptation through steering.

\begin{figure*}[t]
    \centering
    \includegraphics[width=0.96\textwidth]{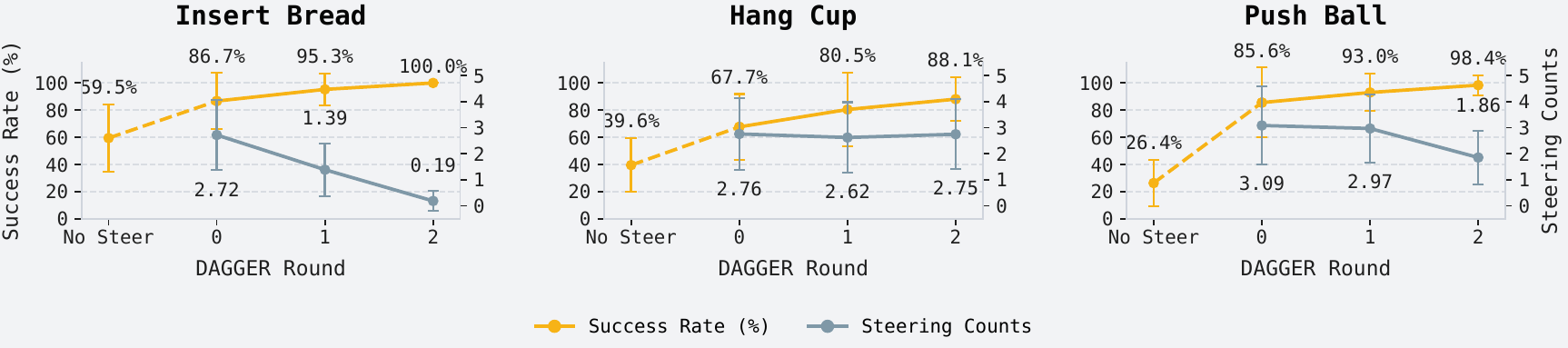}
    \caption{
    Main DAgger results with $\pi_{0.5}$ across three real-world tasks.
    Round~0 denotes the original policy before DAgger fine-tuning.
    Across the Insert Bread, Hang Cup, and Push Ball tasks, iterative training with RoboPrompt-guided corrections improves success rate, while the required steering counts are reduced or remain comparable.
    }
    \label{fig:dagger_main}
\end{figure*}

As shown in Fig.~\ref{fig:dagger_main}, the improvement is significant across different tasks. Each DAgger round increases the task success rate while reducing the steering counts required in the next round, indicating that a policy with moderate initial performance and frequent human intervention can gradually transition toward fully autonomous execution within only a few DAgger iterations.

\begin{figure*}[t]
    \centering
    \includegraphics[width=0.96\textwidth]{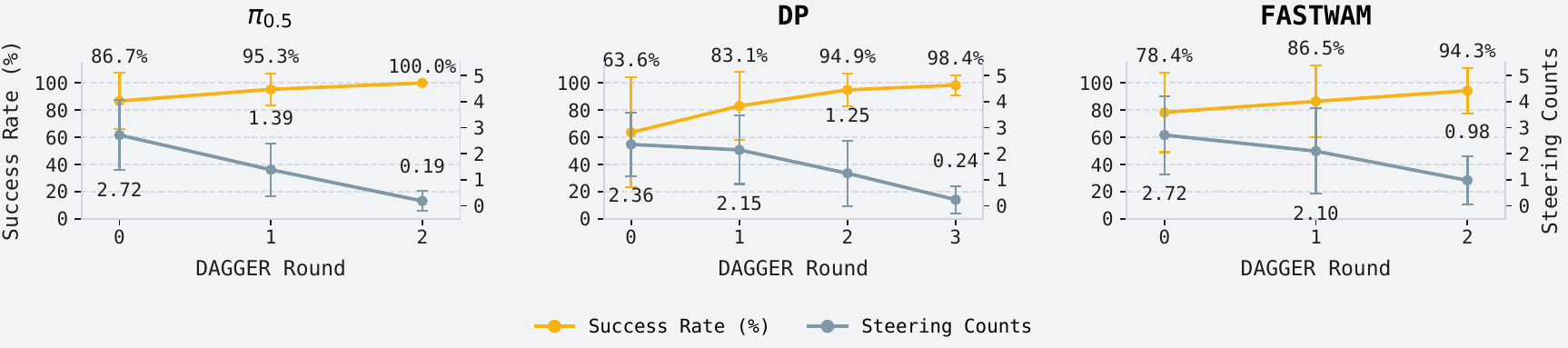}
    \caption{
    DAgger performance after replacing the low-level policy.
    Round~0 denotes the original base policy.
    With additional DAgger iterations, task progress increases while the average steering counts decrease across DP, $\pi_{0.5}$, and FastWAM.
    }
    \label{fig:dagger}
\end{figure*}

We next evaluate DP and FastWAM with the same DAgger protocol on the bread task and observe the same trend, as shown in Fig.~\ref{fig:dagger}: additional DAgger rounds improve task progress while reducing the amount of steering required during deployment. Together, these results suggest that RoboPrompt makes it easier for users to correct policy behavior through sparse, intuitive guidance while retaining the policy improvement ability of DAgger.

\begin{figure*}[t]
    \centering
    \includegraphics[width=0.96\textwidth]{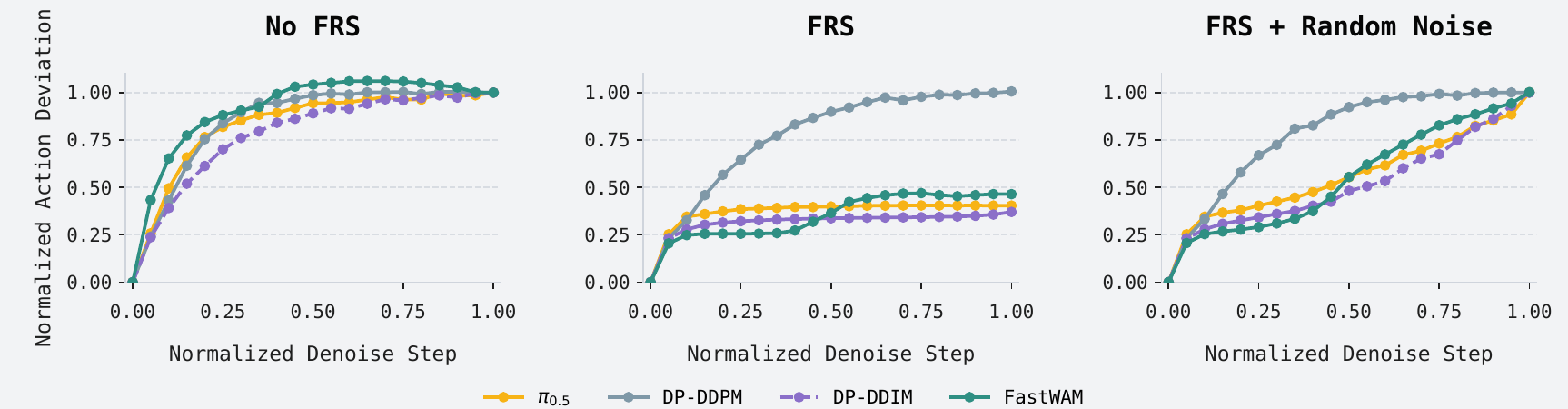}
    \caption{
    \textbf{Denoising dynamics ablation.} Original flow-matching and diffusion-based policies exhibit non-uniform action changes over denoising time. FRS keeps the output closer to the Phase I action, while NA-FRS produces a smoother, approximately linear transition across denoising levels.}
    \label{fig:denoise}
\end{figure*}

\question{How Do Denoising Dynamics Affect RoboPrompt Steering?}
\label{sec:denoise_dynamics}
We study the denoising dynamics of RoboPrompt in Fig.~\ref{fig:denoise}. For both flow-matching and diffusion-based policies, action changes are not uniformly distributed across denoising steps. A small number of denoising steps can already move the action substantially, while the final steps at larger denoising levels are close to convergence and produce only minor changes. This explains why naively increasing the denoising level does not provide a linear control knob between the Phase I action and the base policy output.

NA-FRS changes this behavior by reversing the policy dynamics from the Phase I action before refinement and then injecting controlled random noise. The reverse component keeps the refined action close to the Phase I draft, while the noise component lets the trajectory recover the behavior of pure-noise denoising as the denoising level grows. Together, they produce a smoother, near-linear action transition across denoising levels.

\question{Can first-time users intuitively steer robot policies with RoboPrompt?}
To assess the ease of use of RoboPrompt for first-time users, we recruit 6 volunteers with no prior experience with our system. Following a brief 5-minute introduction, each participant performs 5 trials of the \textit{bread insertion} task using the $\pi_{0.5}$ Round-0 toaster policy shown in Fig.~\ref{fig:dagger_main}. We compare first-time users' average task progress and average steering counts with those of experienced users.

\begin{figure}[t]
\centering
\includegraphics[width=\linewidth]{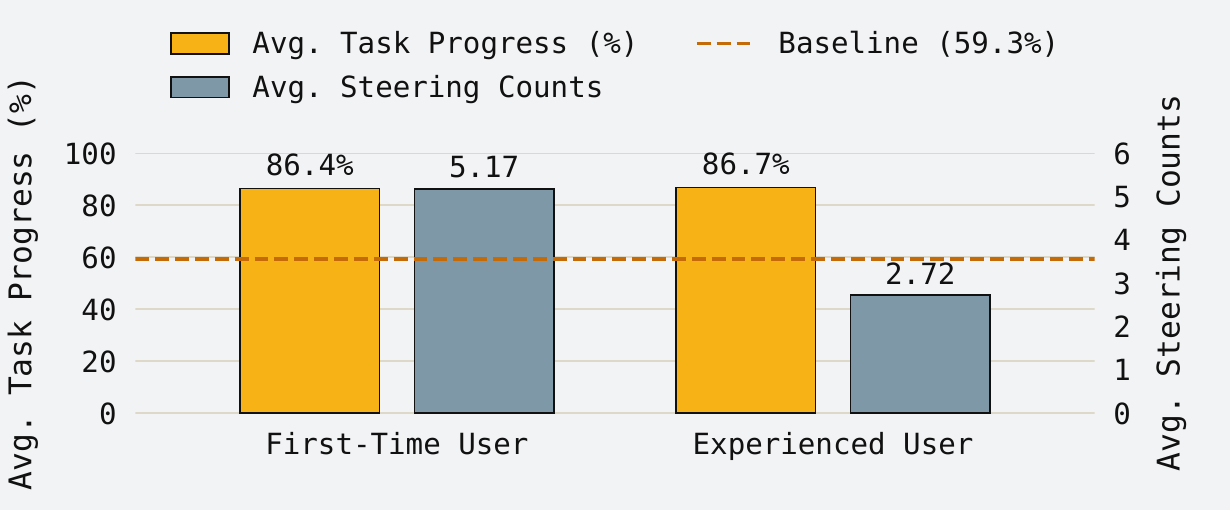}
\caption{Task progress and steering counts for first-time and experienced users on \textit{bread insertion}, using the same policy checkpoint and environment. In baseline, the policy executes autonomously without human steering.}
\label{fig:user_study}
\end{figure}

As shown in Fig.~\ref{fig:user_study}, first-time users achieve task progress comparable to experienced users and substantially above the unsteered baseline, suggesting that RoboPrompt enables new users to intuitively steer robot policies. Experienced users need fewer interventions, as familiarity with the policy helps them anticipate and prevent failures instead of recovering from them.

\question{How does rollout data quality affect policy improvement?}
We compare DAgger training on 20 successful trajectories selected either randomly (unfiltered) or using Time-weighted Optimal Transport (TOT; filtered). Both settings use one training round and are evaluated across 3 tasks.

\begin{figure*}[t]
    \centering
    \includegraphics[width=0.96\textwidth]{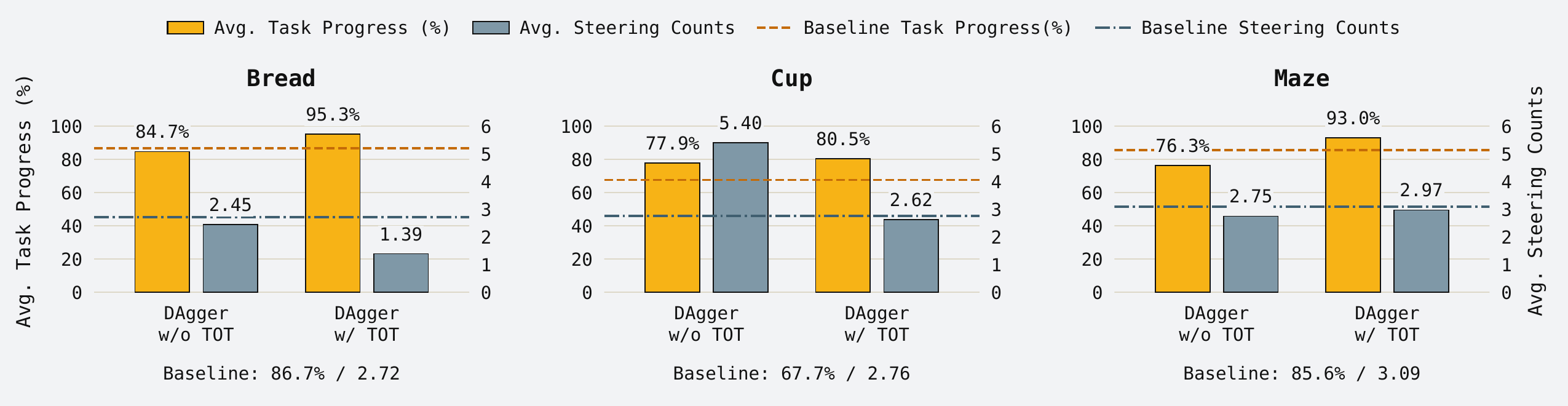}
    \caption{Policy improvement with filtered (w/ TOT) and unfiltered (w/o TOT) rollout data. Each setting uses 20 successful trajectories for one DAgger round. Baselines show task progress and average steering counts before training.}
    \label{fig:dagger_filtering}
\end{figure*}

TOT-filtered data yields higher task progress across all 3 tasks (Fig.~\ref{fig:dagger_filtering}). Unfiltered data increases motion jitter, potentially causing failures or additional interventions; successful execution alone therefore does not ensure training quality. TOT retains smoother trajectories consistent with human intent, improving policy performance and reducing steering counts.

\section{Conclusion and Limitations}
In this work, we present RoboPrompt, a system for steering robot policies with sparse, intuitive human input. By leveraging the properties of diffusion and flow-matching, RoboPrompt enables non-expert users to intuitively guide policy behavior and iteratively improve policy performance using data collected from steered rollouts.

However, while our steering method effectively guides the end-effector pose, it does not yet support controlling dexterous end-effectors, particularly multi-fingered hands. Moreover, our Phase I steering module is not limited to specific robot embodiments or environments. Future work could explore training on larger and more diverse datasets to enable zero-shot steering across a broader range of tasks.
\bibliography{ref}

\end{document}